%% file: main.tex
\documentclass[10pt,twocolumn,letterpaper]{article}

\usepackage[pagenumbers]{iccv} 

\usepackage{graphicx}
\usepackage{amssymb}
\usepackage{booktabs}
\usepackage{colortbl}
\usepackage{pifont}
\definecolor{forestgreen}{RGB}{47, 159, 87}

\usepackage{amsmath}
\usepackage[ruled, linesnumbered]{algorithm2e}
\usepackage{makecell}
\usepackage{color}
\usepackage{caption}
\usepackage{enumitem}


\definecolor{iccvblue}{rgb}{0.21,0.49,0.74}
\usepackage[pagebackref,breaklinks,colorlinks,allcolors=iccvblue]{hyperref}

\def\paperID{11297} 
\def\confName{ICCV}
\def\confYear{2025}

\title{Foresight in Motion: Reinforcing Trajectory Prediction with Reward Heuristics}

\author{Muleilan Pei\textsuperscript{1} \quad 
Shaoshuai Shi\textsuperscript{2}\thanks{Corresponding author.} \quad
Xuesong Chen\textsuperscript{2} \quad
Xu Liu\textsuperscript{3} \quad 
Shaojie Shen\textsuperscript{1} \\
\textsuperscript{1}HKUST \quad
\textsuperscript{2}Voyager Research, Didi Chuxing \quad
\textsuperscript{3}Zhuoyu Technology  \\
\tt\small{\{mpei, eeshaojie\}@ust.hk,}
\tt\small{shaoshuaics@gmail.com}}

\begin{document}
\maketitle
\input{sec/0_abstract}

\input{sec/1_intro}

\input{sec/2_related_work}

\input{sec/3_method}

\input{sec/4_exp}

\input{sec/5_conclusion}

{
    \small
    \bibliographystyle{ieeenat_fullname}
    \bibliography{main}
}

\end{document}

%% file: sec/0_abstract.tex
\begin{abstract}
Motion forecasting for on-road traffic agents presents both a significant challenge and a critical necessity for ensuring safety in autonomous driving systems. In contrast to most existing data-driven approaches that directly predict future trajectories, we rethink this task from a planning perspective, advocating a ``First Reasoning, Then Forecasting” strategy that explicitly incorporates behavior intentions as spatial guidance for trajectory prediction. To achieve this, we introduce an interpretable, reward-driven intention reasoner grounded in a novel query-centric Inverse Reinforcement Learning (IRL) scheme. Our method first encodes traffic agents and scene elements into a unified vectorized representation, then aggregates contextual features through a query-centric paradigm. This enables the derivation of a reward distribution, a compact yet informative representation of the target agent's behavior within the given scene context via IRL. Guided by this reward heuristic, we perform policy rollouts to reason about multiple plausible intentions, providing valuable priors for subsequent trajectory generation. Finally, we develop a hierarchical DETR-like decoder integrated with bidirectional selective state space models to produce accurate future trajectories along with their associated probabilities. Extensive experiments on the large-scale Argoverse and nuScenes motion forecasting datasets demonstrate that our approach significantly enhances trajectory prediction confidence, achieving highly competitive performance relative to state-of-the-art methods.
\end{abstract}

%% file: sec/1_intro.tex
\section{Introduction}
\label{sec:1_intro}

\begin{figure}[t]
    \centering
    \includegraphics[width=0.46\textwidth]{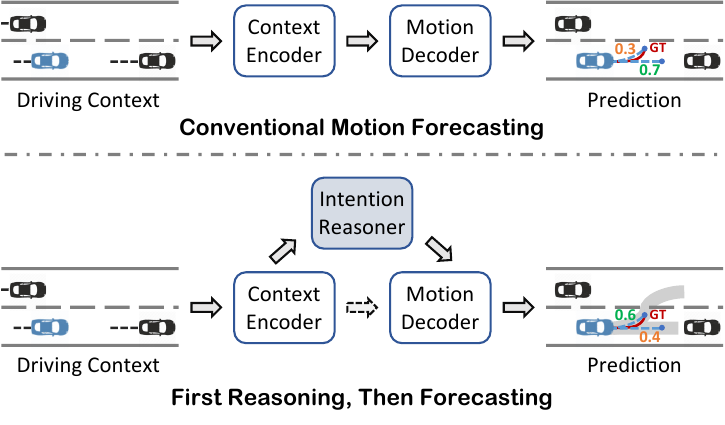}
    \caption{Comparison between the conventional motion forecasting pipeline and ours. Existing approaches mainly predict the trajectory as depicted in the upper framework, while our method follows a ``First Reasoning, Then Forecasting” strategy, illustrated in the lower framework (with the dashed flow indicating an optional step). The blue vehicle represents the target agent to be predicted.}\label{Fig1}
    \vspace{-0.1cm}
\end{figure}

Trajectory prediction is a crucial component of advanced autonomous driving systems, bridging upstream perception with downstream planning modules. Accurately forecasting the movements of surrounding traffic agents requires reasoning about unknown intentions, given the inherent uncertainty and multimodal nature of driving behavior \cite{deo2022multimodal, song2021learning}.

Most existing data-driven motion prediction models \cite{zhou2023query, shi2022motion, gu2021densetnt, lu2022kemp} utilize imitative approaches that either directly regress trajectories or classify endpoints based on data distribution matching from training datasets. However, these methods often lack adequate consideration of driving behavior, limiting interpretability and reliability. While numerous methods achieve strong performance on benchmark metrics, few explicitly reason about future intentions, creating a critical bottleneck in generating interpretable and robust multimodal predictions in real-world applications.

In contrast, human drivers typically maneuver vehicles in a hierarchical manner, making high-level decisions (e.g., lane changes or overtaking) before executing specific motion strategies. Consider the prediction module of the ego vehicle can be viewed as planning on behalf of other agents, under the assumption that road users behave rationally. Despite the intrinsic connection between trajectory prediction and planning, few studies have explored insights from the planning domain \cite{song2020pip}. Motivated by these observations, we pose a key question: \textit{Can trajectory prediction be approached from a planning perspective and enhanced with intention reasoning capabilities?} 

To this end, we advocate a ``First Reasoning, Then Forecasting” strategy, where behavioral intention reasoning provides critical prior guidance to facilitate accurate and confident multimodal motion forecasts. Consider an overtaking scenario as a motivating example: a model capable of explicitly reasoning both overtaking and lane-keeping intents in advance can generate more reliable predictions compared to direct forecasting without reasoning, as shown in \cref{Fig1}. Besides, incorporating longer-term intention reasoning can further enhance prediction confidence (see \cref{table2}).

Nevertheless, due to the inherent complexity of driving scenarios, reasoning about future intentions remains challenging when relying solely on handcrafted rules or predefined planners \cite{song2021learning}. One promising alternative involves utilizing Large Reasoning Models (LRMs), such as OpenAI-o1 \cite{jaech2024openai}, to enable intention reasoning within trajectory predictors. However, their substantial computational demands render them impractical for onboard driving systems. Fortunately, recent advances in LRMs have demonstrated the remarkable reasoning capabilities of Reinforcement Learning (RL) techniques in domains such as mathematics and coding \cite{guo2025deepseek}, prompting an intriguing question: \textit{Can RL-based paradigms be leveraged to reason about an agent’s future intentions in trajectory prediction?}

In this regard, we explore the feasibility of applying RL paradigms to model the behavior reasoning of agents within autonomous driving scenarios. We formulate the task as a Markov Decision Process (MDP) and define the target agent’s intentions accordingly. To balance performance and computational efficiency, we construct a grid-level graph to represent the scene layout, where the intention is defined as a sequence of decisions over a discrete grid world, analogous to a \textit{plan} in the traditional context of RL \cite{sutton1998reinforcement, deo2020trajectory}. The intention sequence is referred to as the Grid-based Reasoning Traversal (GRT) in our paper. However, a fundamental challenge of adopting RL for trajectory prediction lies in modeling the reward, as agent intentions remain unknown.

To overcome this challenge, we propose a reward-driven intention reasoner grounded in Maximum Entropy Inverse Reinforcement Learning (MaxEnt IRL) \cite{ziebart2008maximum}. This framework first learns the agent-specific reward distribution from demonstrations and associated driving contexts via IRL. The learned reward serves as a compact representation \cite{ridel2020scene}, capturing observed behaviors and underlying intentions of the agent. Utilizing these inferred rewards as heuristics, we then perform policy rollouts to sample multiple plausible GRTs and extract their corresponding intention-informed features, providing prior guidance for trajectory prediction and thereby improving prediction accuracy and confidence.

Additionally, to further enhance feature extraction from scene contexts, we introduce a novel Query-centric IRL framework, termed QIRL, which integrates IRL with a query-based encoding mechanism. QIRL efficiently and flexibly aggregates vectorized scene context features into spatial grid-like tokens, facilitating structured reasoning. With this dense grid representation, we augment our model with an auxiliary Occupancy Grid Map (OGM) prediction head, which enables dense predictions of future spatial-temporal occupancies for each agent in the scene. This auxiliary task effectively enhances the feature fusion process by capturing future interactions between agents, improving the overall prediction performance (see \cref{ablation_4}).

Finally, to fully exploit features provided by the intention reasoner, we develop a hierarchical DETR-like trajectory decoder. An anchor-free trajectory token initially generates proposals conditioned on GRT-derived features, which then act as initialized anchors for final trajectory decoding. Given the inherent sequential nature of trajectory states and recent advancements in selective state-space models (Mamba) for long-term, structured dynamic modeling \cite{gu2023mamba, zhu2024vision}, we incorporate a bidirectional variant, Bi-Mamba, to effectively capture the sequential dependencies of trajectory states. This enhancement significantly improves both prediction accuracy and confidence (see \cref{ablation_3}).

In summary, our main contributions are as follows:

\begin{enumerate}[leftmargin=0.5cm, label=(\arabic*)]
    \item We introduce a ``First Reasoning, Then Forecasting'' strategy, rethinking trajectory prediction tasks from a planning perspective.

    \item We propose a novel reward-driven intention reasoner for motion forecasting, with a QIRL module that integrates the MaxEnt IRL paradigm and vectorized context representation in a query-centric framework.

    \item We develop a hierarchical DETR-like decoder with bidirectional selective state-space models (Bi-Mamba) to improve prediction accuracy and confidence.

    \item Our approach significantly boosts the prediction confidence and achieves highly competitive performance on the Argoverse and nuScenes motion forecasting benchmarks, outperforming other state-of-the-art models.
\end{enumerate}

%% file: sec/2_related_work.tex
\section{Related Work}
\label{sec:2_related_work}

\begin{figure*}[t]
    \centering
    \includegraphics[width=0.98 \textwidth]{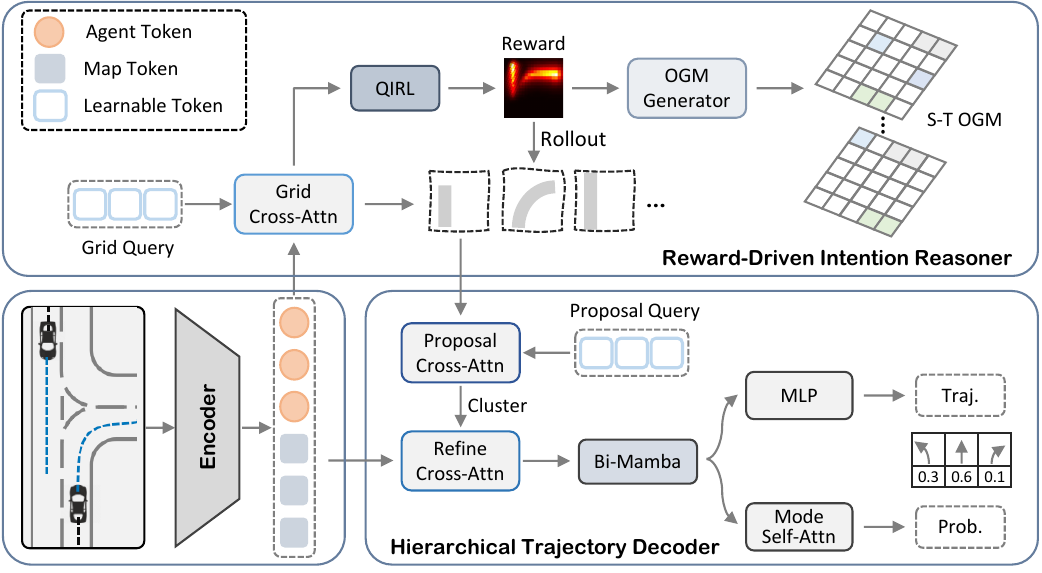}
    \caption{Overview of the proposed framework. Our approach integrates a reward-driven intention reasoner into an encoder-decoder motion forecasting structure. Scene context features are initially extracted using simple encoders and subsequently aggregated into grid queries. The proposed QIRL module generates reward distributions that reason about multiple plausible intention sequences as priors. These sampled tokens are then utilized to generate trajectory proposals, which, after clustering and refinement, yield multimodal future trajectories with associated confidences through a hierarchical DETR-like trajectory decoder enhanced by Bi-Mamba. Additionally, an auxiliary Spatial-Temporal OGM prediction head is incorporated to further improve feature fusion.} \label{Fig2}
    \vspace{-0.4cm}
\end{figure*}

\noindent \textbf{Trajectory Prediction} for autonomous driving has been a subject of investigation for decades. Early efforts in this domain primarily relied on handcrafted rule-based or physics-based methods, which struggled to handle complex scenarios and lacked the capability for long-term predictions \cite{mozaffari2020deep}. Recent approaches have transitioned to learning-based frameworks that leverage deep neural networks to encode agents' historical motion profiles while integrating the topological and semantic information \cite{pei2025sept} of High-Definition (HD) maps. These maps are represented in either rasterized \cite{cui2019multimodal, chai2019multipath} or vectorized \cite{liang2020learning, gao2020vectornet} formats. Rasterized representations typically use Bird's-Eye-View (BEV) images as input, while vectorized representations rely on agent and map polylines as input. Convolutional Neural Networks (CNNs) and Graph Neural Networks (GNNs) are widely employed as feature extractors for these formats, playing a crucial role in encoding scene context.
More recently, Transformer-based architectures have garnered significant attention for feature extraction and fusion, owing to their ability to improve overall prediction performance \cite{ngiam2021scene, shi2022motion, zhou2023query}. Following this trend, our work adopts vectorized representations and leverages a query-centric Transformer encoder-decoder structure for feature aggregation and trajectory generation.

Despite these advancements, challenges persist in making trajectory prediction robust to out-of-distribution scenarios and generalizable to unseen environments. Our work addresses these gaps by rethinking trajectory prediction from a planning perspective, introducing a reward-driven intention reasoner to provide behavior guidance and contextually rich priors for advancing trajectory prediction.

\noindent \textbf{Reward} is a foundational concept in both planning and RL, serving as a guiding signal that shapes an agent’s behavior and decision-making process \cite{pei2020quadruped}. In planning, rewards are often designed to align with high-level objectives, such as reaching a goal while avoiding obstacles \cite{pei2021improved}. Typically, reward functions are either handcrafted or shaped through hierarchical frameworks, where high-level planners provide strategic guidance to lower-level controllers. Research on reward shaping has demonstrated that modifying the reward structure to emphasize specific behaviors or milestones can accelerate learning and improve policy robustness. In RL, the reward function plays a central role, defining the agent’s objective and guiding it to perform actions that maximize cumulative rewards over time. The design of reward functions is crucial in both planning and RL; however, crafting effective rewards for complex tasks, such as autonomous driving, is exceptionally challenging. To address this challenge, Inverse RL (IRL) \cite{russell1998learning, ng2000algorithms} was introduced. IRL focuses on inferring a reward function from observed expert demonstrations, making it particularly valuable in scenarios where directly defining a reward function is infeasible. For example, MaxEnt IRL \cite{ziebart2008maximum} has been widely applied to learn reward functions that capture the underlying intentions of expert behaviors \cite{wu2020efficient}, enabling agents to replicate nuanced, human-like decision-making in planning tasks \cite{huang2023conditional}.

Despite their utility, existing efficient IRL algorithms are typically tailored for structured and grid-like environments \cite{fernando2020deep}, limiting their flexibility in more complex domains. To overcome this limitation, we propose a novel query-centric framework that enhances the applicability and flexibility of MaxEnt IRL in our reward-driven intention reasoner. By leveraging this paradigm, our approach provides valuable reward heuristics that effectively reason about intentions of future behavior, providing informative priors for addressing the complexities inherent in motion forecasting tasks.

%% file: sec/3_method.tex
\section{Methodology}
\label{sec:3_method}

\subsection{Problem Formulation}
The goal of the standard trajectory prediction task is to forecast the future positions of a target agent over a time horizon \( T_f \), given the driving context. We adopt vectorized representations for the scene input, including the historical observed states \( A \in \mathbb{R}^{N_a \times T_h \times C_a} \), where \( N_a \) denotes the number of agents in the scene, \( T_h \) represents the number of past timestamps, and \( C_a \) captures motion profiles such as positions, velocities, headings, etc, along with HD map information  \( M \in \mathbb{R}^{N_m \times N_s \times C_m} \), where \( N_m \) and \( N_s \) correspond to the number of lane centerlines and lane segments, respectively, and \( C_m \) indicates the associated lane attributes.

Our approach employs a target-centric coordinate system in which all input elements are normalized to the current state of the target agent through translation and rotation operations. Given the inherent uncertainty of motion intentions, the predictor is tasked with providing \( K \) future trajectories \( Y \in \mathbb{R}^{K \times T_f \times 2} \), along with the corresponding probabilities \( p \in \mathbb{R}^{K \times 1} \).

\subsection{Framework Overview}
As shown in \cref{Fig2}, our motion forecasting approach adopts an encoder-decoder structure comprising a query-centric scene context encoder, a Mamba-enhanced hierarchical trajectory decoder, and a reward-driven intention reasoner. First, we represent the driving context in a vectorized format and utilize the agent and map encoder to extract scene features. These fused features are then aggregated into spatial grid tokens through cross-attention mechanisms. Subsequently, in the QIRL module, a grid-based MaxEnt IRL algorithm is leveraged to infer the reward distribution, thereby reasoning about multiple plausible intention sequences (i.e., GRTs) over the 2-D grid map through policy rollouts. Additionally, a dense prediction head for the Spatial-Temporal Occupancy Grid Map (S-T OGM) is incorporated to model future interactions among agents. Finally, we introduce a hierarchical DETR-like \cite{carion2020end} trajectory decoder that generates trajectory proposals, which are further clustered and refined to produce multimodal future trajectories enhanced by the Bi-Mamba architecture.

\subsection{Query-Centric Context Encoding}
Given the vectorized agent representations \( A \) and map representations \( M \), we first tokenize them into separate feature sets. Specifically, we use an agent encoder, a simple 1-D CNN model \cite{liang2020learning}, to obtain the agent features \( F_a \in \mathbb{R}^{N_a \times C} \). For the map encoder, we adopt a PointNet-like network \cite{qi2017pointnet, gao2020vectornet} to extract static map features as \( F_m \in \mathbb{R}^{N_m \times C} \). These resulting agent and map features are then concatenated to form context tokens \( F_c \in \mathbb{R}^{(N_a + N_m) \times C} \), followed by a self-attention block to enhance feature fusion.

Since the reasoning process relies on a grid-level graph representation, we introduce learnable grid-shaped queries \( Q_G \in \mathbb{R}^{H \times W \times C} \) to integrate the scene features, where \( H \) and \( W \) define the spatial dimensions in the BEV plane. Each query \( Q_G^{s_i} \) at grid position \( s_i = (x_i, y_i) \) corresponds to a specific region in the real world with a resolution \( d \). We then use the flattened grid queries with 2-D spatial learnable relative positional embeddings to aggregate the context tokens via a cross-attention mechanism.

\subsection{Reward-Driven Intention Reasoning} \label{S-2}
Given the grid tokens updated with context features, we first generate the reward distribution through our QIRL framework, which adapts the traditional grid-based MaxEnt IRL algorithm \cite{ziebart2008maximum} in a query-centric paradigm. The MaxEnt IRL is typically defined as a finite MDP model, encompassing a state space, an action space, and a transition model. Its objective is to recover the reward distribution of the environment to yield a policy that mimics expert demonstrations by maximizing the log-likelihood of the demonstration data while adhering to the maximum entropy principle. Demonstrations consist of discrete state sequences, with rewards generally formulated as a combination of environmental features. The learning process involves an inner-loop forward RL process within each reward iteration until the loss \( \mathcal{L}_{IRL} \) converges. More technical details on the MaxEnt IRL can be found in Appx. 6.1 and \cite{pei2025goirl, wulfmeier2017large}. 

\noindent \textbf{QIRL.} In our QIRL framework, each grid \( s_i \) acts as a state, with its corresponding query \( Q_G^{s_i} \in \mathbb{R}^{1 \times C} \) representing contextual features. We aggregate features from grid tokens with a stack of \( 1 \times 1 \) CNN layers to establish a nonlinear mapping from driving context to the reward \( R \in \mathbb{R}^{H \times W \times 1} \). Future trajectories are quantized to a resolution \( d \) to form expert demonstration states, which can also include paths for capturing long-term information, if available. Subsequently, the MaxEnt IRL algorithm is applied to derive the converged reward distribution alongside an optimal policy. 

Rollouts are then performed based on the policy induced by the reward heuristic. We execute \( L \) rollouts over the grid map in parallel, producing multiple plausible GRTs as sequences of intentions, \( \Upsilon \in \mathbb{R}^{L \times \mathcal{H} \times 2} \), where \( \mathcal{H} \) denotes the planning horizon. To better capture the multimodal future distribution, we set \( L \gg K \). The grid tokens are then extracted according to the sampled GRT: for each position \( (x_i, y_i) \) associated with grid cell state \( s_i \) in the sampled GRT, the corresponding grid token \( Q_G^{s_i} \) is sequentially selected over \( \mathcal{H} \) steps. These grid tokens form the reasoning tokens \( Q_G^\Upsilon \in \mathbb{R}^{L \times \mathcal{H} \times C} \). The GRT positions \( \Upsilon \) and associated reasoning tokens \( Q_G^\Upsilon \) serve as informative behavior intention priors for guiding subsequent motion forecasting.

\noindent \textbf{Auxiliary S-T OGM Prediction Head.} Leveraging the grid-shaped dense representations, we introduce an auxiliary S-T OGM prediction head to model future interactions among agents, thereby enhancing the fusion and aggregation of scene context features. We represent the occupancy map in binary form, where grid cells occupied in the BEV at future timestamps are set to 1, and unoccupied cells are set to 0. Our OGM generator takes the fused grid tokens \( Q_G \) and reward \( R \) as input and produces OGMs over \( T_f \) future timestamps using a U-Net-like architecture \cite{ronneberger2015u}.

\begin{figure}[t]
    \centering
    \includegraphics[width=0.36\textwidth]{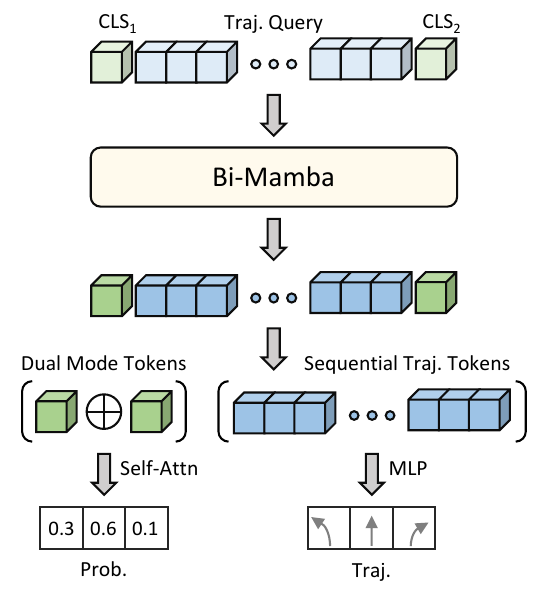}
    \caption{The process of Bi-Mamba-enhanced decoding.}  \label{Fig3}
    \vspace{-0.4cm}
\end{figure}

\subsection{Mamba-Enhanced Trajectory Decoding}
Given \( L \) plausible reasoning priors, we first use a DETR-like trajectory generator to produce \( L \) trajectories as proposals. We encode the GRT positions \( \Upsilon \) and reasoning tokens \( Q_G^\Upsilon \) separately through simple MLP blocks, then concatenate and process them via an MLP-based feature fusion network to form the final reasoning tokens \( Q_\Upsilon \in \mathbb{R}^{L \times \mathcal{H} \times C} \). Next, we introduce an anchor-free learnable trajectory proposal query \( Q_P \in \mathbb{R}^{L \times T_f \times C} \) to cross-attend to the prior features \( Q_\Upsilon \) from the intention reasoner. This proposal query is then decoded into \( L \) trajectory proposals by a regression head with an MLP block. We apply the K-means algorithm to cluster these proposals into \( K \) multimodal trajectory proposals \( \overline{Y} \in \mathbb{R}^{K \times T_f \times 2} \). Subsequently, we use an anchor-based trajectory refinement, as in many existing motion predictors \cite{zhang2024simpl}, to further enhance trajectory query prediction performance. Each trajectory proposal, acting as an explicit anchor prior, is re-encoded into the trajectory query \( Q_T \in \mathbb{R}^{K \times T_f \times C} \), which retrieves the original context features through a DETR-like architecture similar to that used for trajectory proposal generation. This hierarchical anchor-free proposal generation, together with the anchor-based refinement process, results in a trajectory query that integrates both reward-driven intentions and detailed scene context.

\noindent \textbf{Bi-Mamba Decoder.} Since the trajectory token \( Q_T \) contains significant sequential properties in both time and spatial domains, we employ a selective state-space model to capture the coupled relationships within the series of trajectory queries, inspired by the recent success of the Mamba architecture in sequential modeling. Specifically, we adopt a Bi-Mamba model to process the trajectory token, leveraging its bidirectional scanning mechanism for more comprehensive information capture. In this Bi-Mamba-enhanced decoding process, we predict trajectory offsets \( \Delta Y \in \mathbb{R}^{K \times T_f \times 2} \) and the probability of each hypothesis \( p \).

To better leverage the bidirectional capabilities of the Bi-Mamba structure, we design a learnable dual mode query \( Q_M \in \mathbb{R}^{K \times 2 \times C} \), containing two classification (\( \text{CLS} \)) tokens. These tokens, \( \text{CLS}_1 \) and \( \text{CLS}_2 \), are appended before and after the trajectory query \( Q_T \), as illustrated in \cref{Fig3}. The tokens aggregate backward and forward features, respectively, offering a more comprehensive fusion compared to a unidirectional Mamba with a single classification token, as validated in our ablation results (see \cref{ablation_4}). After the Bi-Mamba process, the two \( \text{CLS} \) tokens are combined through element-wise addition for feature fusion. A mode self-attention module then enables interaction among modes, further enhancing the multimodality of predictions. 

Finally, the mode token undergoes classification with a softmax function to generate probabilities, and the sequential trajectory tokens are decoded with a regression head to produce trajectory offsets. More details are in Appx. 6.2.

The final predicted trajectory \( Y \) is obtained by summing the trajectory proposal \( \overline{Y} \) and its associated offset \( \Delta Y \), as follows:
\begin{equation}
Y = \overline{Y} + \Delta Y.
\end{equation}

\subsection{Training Objectives}
Our entire pipeline involves multiple training objectives. The reward-driven intention reasoner includes two subtask objectives: the QIRL and OGM generator. The QIRL objective employs the loss \( \mathcal{L}_{IRL} \) as discussed in \cref{S-2}, while the OGM generator, denoted as \( \mathcal{L}_{OGM} \), uses a focal Binary Cross-Entropy (BCE) loss. 

For the trajectory decoder, the training objectives include a regression loss \( \mathcal{L}_{REG} \) and a classification loss \( \mathcal{L}_{CLS} \). To optimize trajectory regression, we apply Huber loss to both the trajectory proposals and the refined trajectories. Additionally, to address the mode collapse issue, we employ a winner-takes-all strategy, commonly used in similar works, where only the candidate with the minimum displacement error is selected for backpropagation. For mode classification, we adopt a max-margin loss following \cite{liang2020learning}.

The overall loss \( \mathcal{L} \) integrates these components and can be optimized in an end-to-end fashion:
\begin{equation}
\mathcal{L} = \mathcal{L}_{IRL} + \alpha \mathcal{L}_{OGM} +
\beta \mathcal{L}_{REG} + \gamma \mathcal{L}_{CLS},
\end{equation}
where \( \alpha \), \( \beta \), and \( \gamma \) are hyperparameters to balance each training objective. More details can be found in Appx. 6.3.

\begin{table*}[t]
	\centering
        \resizebox{0.72\textwidth}{!}{
	\begin{tabular}{l|cccc >{\columncolor[gray]{0.9}}c}
	\toprule
	Method           & MR$_{6}\downarrow$  & minADE$_{6}\downarrow$      & minFDE$_{6}\downarrow$  & brier-minFDE$_{6}\downarrow$ & Brier score $\downarrow$ \\
	\midrule
    DenseTNT~\cite{gu2021densetnt}   & 0.1258  & 0.8817 & 1.2815 & 1.9759 & 0.6944 \\
    HiVT~\cite{zhou2022hivt}    & 0.1267 & \textbf{0.7735} & \textbf{1.1693} & \underline{1.8422} & 0.6729 \\
    SceneTransformer~\cite{ngiam2021scene}      & \underline{0.1255} & \underline{0.8026} & 1.2321 & 1.8868 & 0.6547 \\
    DSP~\cite{zhang2022trajectory} &0.1303 &0.8194 &1.2186 &1.8584 & \underline{0.6398} \\
    FiM (Ours)          & \textbf{0.1250} & 0.8296 & \underline{1.2048} & \textbf{1.8266} & \textbf{0.6218}  \\
    
    \midrule
    SIMPL~\cite{zhang2024simpl}          & 0.1165 & 0.7693 & 1.1545 & 1.7469 & 0.5924 \\
    HPNet~\cite{tang2024hpnet}    & \textbf{0.1075} & \textbf{0.7478}  & \textbf{1.0856} & \textbf{1.6768} & 0.5912 \\
    Wayformer~\cite{nayakanti2022wayformer}      & 0.1186 & \underline{0.7676} & 1.1616 & 1.7408 & 0.5792 \\
    MultiPath++~\cite{varadarajan2022multipath++}	    & 0.1324 & 0.7897 & 1.2144 & 1.7932 & \underline{0.5788} \\
    FiM (Ours)            & \underline{0.1087} & 0.7795 & \underline{1.1199} & \underline{1.6931} & \textbf{0.5732}  \\

  \bottomrule
\end{tabular}
}
\caption{Performance comparison on the \textbf{Argoverse 1} motion forecasting leaderboard. The upper group lists results of single models, while the lower group lists those with ensemble methods. The best and the second-best results are in \textbf{bold} and \underline{underlined}, respectively. \label{table1}}
\vspace{-0.2cm}
\end{table*}

%% file: sec/4_exp.tex
\section{Experiments and Results}
\label{sec:4_exp}

\begin{table}[!tbp]
	\centering  
    \resizebox{0.478\textwidth}{!}{
	\begin{tabular}{l|cc>{\columncolor[gray]{0.9}}c}
	\toprule
	Method  & minFDE$_{6}\downarrow$ & brier-minFDE$_{6}\downarrow$ & Brier score $\downarrow$   \\
	\midrule
    
    QCNet~\cite{zhou2023query}  & 0.551 & 1.180 & 0.629 \\
    DeMo~\cite{zhang2024decoupling}   & 0.543          & 1.169           & 0.626   \\
    FiM w/ GRT-S   & \underline{0.529} & 1.147 & 0.617 \\
    FiM w/ GRT-M  & 0.530 & \underline{1.134} & \underline{0.604} \\
    FiM w/ GRT-L  & \textbf{0.528} & \textbf{1.131} & \textbf{0.603} \\
  \bottomrule
\end{tabular}
}
\caption{Performance comparison on the customized \textbf{Argoverse 2} validation set. Results for QCNet and DeMo are produced using their official implementations.}  
\label{table2}
\vspace{-0.2cm}
\end{table}

\subsection{Experimental Settings}
\noindent \textbf{Datasets.} We use various large-scale public datasets, Argoverse 1 \cite{chang2019argoverse}, Argoverse 2 \cite{wilson2023argoverse}, and nuScenes \cite{caesar2020nuscenes}, for training and evaluating our approach. All datasets provide rich HD map information. The Argoverse 1 dataset includes over 1,000 hours of driving data from Miami and Pittsburgh, with approximately 206k, 39k, and 78k sequences for training, validation, and testing, respectively. Each sequence consists of 50 timestamps sampled at 10 Hz. The task is to forecast the next 30 positions given the past 20 observed states (i.e., \( T_h = 20 \), \( T_f = 30 \)). The Argoverse 2 dataset collects 250k scenarios from six cities, with about 200k and 25k sequences for training and validation, respectively. Each sequence comprises 110 timestamps sampled at 10 Hz, with \( T_h = 50 \) for observations and \( T_f = 60 \) for future predictions. The nuScenes dataset, collected in Boston and Singapore, presents dense traffic and challenging driving scenarios. It includes around 32k, 8k, and 9k sequences for training, validation, and testing, respectively, with each sequence containing 16 timestamps at a 2 Hz sampling frequency. Herein, the task is to predict 12 future positions given the previous 4 observed states (i.e., \( T_h = 4 \), \( T_f = 12 \)).

\noindent \textbf{Metrics.} We follow the standard metrics to evaluate prediction performance, including Miss Rate (\( \text{MR}_K \)), minimum Average Displacement Error (\( \text{minADE}_K \)), minimum Final Displacement Error (\( \text{minFDE}_K \)), Brier minimum Final Displacement Error (\( \text{brier-minFDE}_K \)), and Brier score. Specifically, \( \text{MR}_K \) calculates the proportion of scenarios where none of the \( K \) predicted trajectories are within 2.0 meters of the Ground Truth (GT) endpoint. \( \text{minADE}_K \) measures the average \( \ell_2 \) distance between the closest predicted trajectory and the GT, while \( \text{minFDE}_K \) focuses solely on endpoint error. Furthermore, the Brier score and \( \text{brier-minFDE}_K \), introduced by Argoverse, incorporate prediction confidence into performance evaluation. \( \text{brier-minFDE}_K \) can be obtained by adding the Brier score \( (1.0 - p)^2 \) to \( \text{minFDE}_K \), where \( p \) is the probability of the best forecast. Note that the Brier score is particularly valuable in real-world applications, as accurate prediction confidence and uncertainty are crucial for safe and efficient decision-making and planning.

\begin{table}[!t]
	\centering
        \resizebox{0.478\textwidth}{!}{
	\begin{tabular}{l|>{\columncolor[gray]{0.9}}c ccc c}
	\toprule
	Method   & minADE$_{5}\downarrow$ & MR$_{5}\downarrow$ & minADE$_{10}\downarrow$   & MR$_{10}\downarrow$   \\
	\midrule
    P2T~\cite{deo2020trajectory}	& 1.45 & 0.64 & 1.16  & 0.46  \\
    THOMAS~\cite{gilles2021thomas}	& 1.33 & 0.55  & 1.04 & 0.42  \\
    PGP~\cite{deo2022multimodal}    & 1.27 & 0.52 & 0.94  & 0.34   \\
    DeMo~\cite{zhang2024decoupling}   & 1.22 & 0.43 & 0.89   & 0.34   \\
    MacFormer~\cite{feng2023macformer}     & 1.21 & 0.57 & 0.89   & 0.33  \\
    Goal-LBP~\cite{yao2023goal}       & 1.02 & \underline{0.32} & 0.93   & \underline{0.27}   \\
    UniTraj~\cite{feng2024unitraj} & \underline{0.96} & 0.43 & \underline{0.84}   & 0.41 \\
    FiM (Ours)   &  \textbf{0.88} & \textbf{0.31} & \textbf{0.78}   & \textbf{0.23}    \\
  \bottomrule   
\end{tabular}
}
\caption{Performance comparison on the \textbf{nuScenes} prediction leaderboard. minADE$_{5}$ is the official ranking metric.} 
\label{table3}
\vspace{-0.2cm}
\end{table}

\subsection{Comparison with State of the Art}
We conduct a comprehensive comparison of our approach against state-of-the-art methods on the Argoverse 1, Argoverse 2, and nuScenes motion forecasting datasets. We refer to our method as FiM (Foresight in Motion) for brevity.

\noindent \textbf{Argoverse 1.}
Quantitative results on the Argoverse 1 test split are presented in \cref{table1}. We compare our FiM against several representative published methods evaluated on this challenging benchmark.  According to the single-model results (upper group), FiM achieves highly competitive performance compared with strong baselines, including direct trajectory forecasting models such as HiVT \cite{zhou2022hivt} and SceneTransformer \cite{ngiam2021scene}, as well as goal-based models like DSP \cite{zhang2022trajectory} and DenseTNT \cite{gu2021densetnt}. FiM particularly excels in terms of the Brier score, brier-minFDE$_6$, and MR${_6}$, highlighting its robust predictive capabilities.

We also apply model ensembling techniques to further enhance overall performance. The ensemble result (lower group) demonstrates a substantial performance improvement, indicating the significant potential and upper-bound capability of our proposed framework. Compared to other leading published methods such as HPNet \cite{tang2024hpnet} and Wayformer \cite{nayakanti2022wayformer}, FiM consistently achieves competitive performance across various evaluation metrics, notably excelling in the Brier score. This result emphasizes that our reasoning-enhanced predictor effectively generates predictions with greater reliability and confidence, as intended.

\begin{table}[!tbp]
	\centering
    \resizebox{0.45\textwidth}{!}{
	\begin{tabular}{c|ccc}
	\toprule
	Method  & minFDE$_{6}\downarrow$ & brier-minFDE$_{6}\downarrow$ & Brier score $\downarrow$   \\
	\midrule
    Vanilla   & 2.185          & 2.879           & 0.694   \\
    C.A.  & 1.490 & 2.132 & 0.642 \\
     Ours   & \textbf{1.008} & \textbf{1.602} & \textbf{0.594} \\
  \bottomrule
\end{tabular}
}
\caption{Ablation on the reward-driven reasoning strategy.}
\label{ablation_1}
\end{table}

\begin{table}[!tbp]
	\centering
    \resizebox{0.45\textwidth}{!}{
	\begin{tabular}{cc|ccc}
	\toprule
	OGM & Refine  & minFDE$_{6}\downarrow$ & brier-minFDE$_{6}\downarrow$ & Brier score $\downarrow$  \\
	\midrule
    \ding{51}   &   & 1.059          & 1.670           & 0.611   \\
     &  \ding{51}   & 1.175          & 1.801          & 0.626   \\
     \ding{51} & \ding{51}  & \textbf{1.008} & \textbf{1.602} & \textbf{0.594} \\
  \bottomrule
\end{tabular}
}
\caption{Ablation on the OGM and refinement modules.}
\label{ablation_2}
\vspace{-0.4cm}
\end{table}

\noindent \textbf{Argoverse 2.}
To further validate the effectiveness of our intention reasoning strategy, we introduce a customized evaluation benchmark built on top of the Argoverse 2 validation split. Specifically, the task requires forecasting the first 30 future positions, while during training, models have access to the subsequent 30 positions exclusively as auxiliary intention supervision. Notably, all models are constrained to training for trajectory generation using supervision strictly aligned with the first 30 future positions. This setup simulates practical applications where long-term paths are available for intention learning. Given that the proposed QIRL module is agnostic to the supervision format, whether trajectories or paths, we develop three model variants that incorporate different horizons of future supervision for the GRT training. These variants, denoted as GRT-S, GRT-M, and GRT-L, correspond to reasoning modules trained with 30, 45, and 60 future timestamps, respectively. 

We compare our FiM against the two top-performing open-source models on the Argoverse 2 leaderboard, DeMo \cite{zhang2024decoupling} and QCNet \cite{zhou2023query}. As shown in \cref{table2}, all FiM variants surpass these two strong baselines, demonstrating substantial gains attributed to the intention reasoning module. Moreover, the results further indicate that longer-term intention supervision significantly enhances prediction confidence, facilitating more reliable trajectory forecasts.

\noindent \textbf{nuScenes.}
We also evaluate FiM on the nuScenes dataset, as reported in \cref{table3}. Our model delivers top-tier performance on this prediction benchmark, outperforming all current entries on the leaderboard and further validating the robustness and advanced capabilities of our proposed framework in addressing complex motion forecasting challenges.

\subsection{Ablation Study}

We perform in-depth ablation studies on the Argoverse validation set to assess the effectiveness of key components in our approach, keeping all experimental settings consistent for a fair comparison.

\begin{table}[!tbp]
	\centering
    \resizebox{0.45\textwidth}{!}{
	\begin{tabular}{ccc|ccc}
	\toprule
	MLP & Bi-Mamba  &  Self-Attn. & brier-minFDE$_{6}\downarrow$ & Brier score $\downarrow$   \\
	\midrule
    \ding{51} &  &     & 1.720     & 0.622  \\
     \ding{51} & \ding{51} &    & 1.649    & 0.605  \\ 
     \ding{51} &  & \ding{51}        & 1.682     & 0.617   \\
    \ding{51} & \ding{51} & \ding{51}   & \textbf{1.602} & \textbf{0.594} \\
  \bottomrule
\end{tabular}
}
\caption{Ablation on components of Mamba-enhanced decoder.}
\label{ablation_3}
\end{table}

\begin{table}[!tbp]
	\centering
    \resizebox{0.45\textwidth}{!}{
	\begin{tabular}{l|ccc}
	\toprule
	Mamba   & minFDE$_{6}\downarrow$ & brier-minFDE$_{6}\downarrow$ & Brier score $\downarrow$  \\
	\midrule
        Unidirectional    & 1.034         & 1.636           & 0.603   \\
     Bidirectional   & \textbf{1.008} & \textbf{1.602} & \textbf{0.594} \\
  \bottomrule
\end{tabular}
}
\caption{Comparison between Uni-Mamba and Bi-Mamba.}
\label{ablation_4}
\end{table}

\begin{table}[!tbp]
	\centering
    \resizebox{0.45\textwidth}{!}{
	\begin{tabular}{c|ccc}
	\toprule
	\# of layers  & minFDE$_{6}\downarrow$ & brier-minFDE$_{6}\downarrow$ & Brier score $\downarrow$   \\
	\midrule
    2   & 1.027          & 1.635           & 0.609   \\
     4  & \textbf{1.008} & \textbf{1.602} & \textbf{0.594} \\
     8   & 1.024          & 1.628          & 0.609   \\
  \bottomrule
\end{tabular}
}
\caption{Effect of Mamba layer depth.}
\label{ablation_5}
\vspace{-0.4cm}
\end{table}

\noindent \textbf{Effects of Reward Heuristics.}
We first examined the effectiveness of the reward-driven intention reasoner by removing the reasoning branch from the pipeline. As shown in \cref{ablation_1}, the performance of the vanilla architecture dropped significantly compared to our full model, underscoring the critical contribution of the reasoning process to overall performance. Additionally, we explored the specific impact of the QIRL module by replacing it with a cross-attention block for feature extraction. The results in \cref{ablation_1} show that our QIRL module outperformed this alternative by a large margin, demonstrating that QIRL can effectively gather essential intention priors and provide informative guidance that benefits subsequent motion forecasting. 

\begin{figure*}[t]
    \centering
    \includegraphics[width=\textwidth]{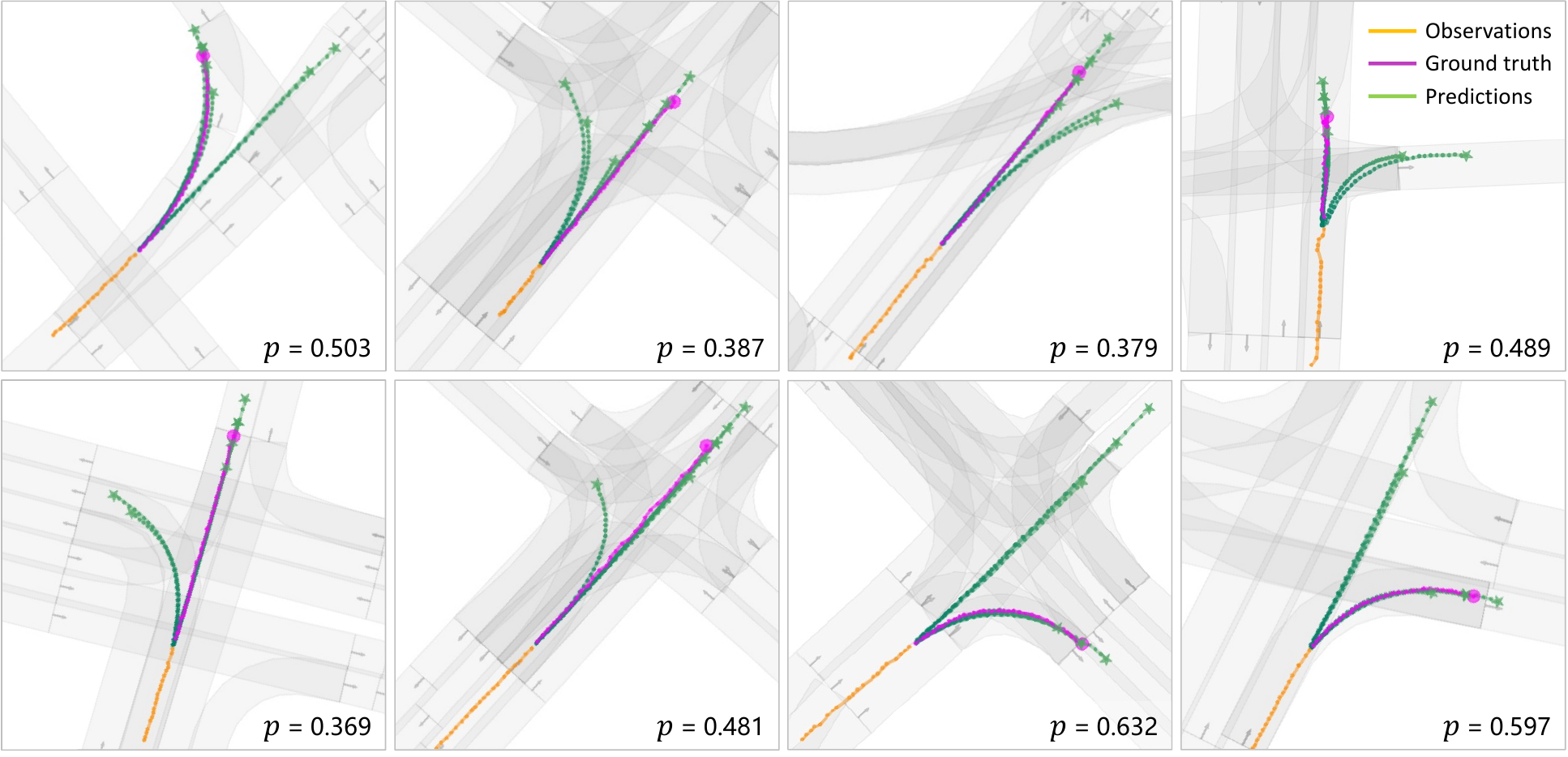}
    \caption{Qualitative results of our model on the Argoverse validation set. The historical trajectory, ground-truth future trajectory, and multimodal predictions are depicted in yellow, magenta, and green, respectively. The lower-right corner showcases the probability associated with the best forecast in terms of the endpoint. Other traffic participants are omitted to better highlight the predictions for the target agent.}\label{Fig4}
    \vspace{-0.4cm}
\end{figure*}

\noindent \textbf{Effects of the OGM \& Refinement.}
We further evaluated the impact of the auxiliary spatial-temporal OGM module and refinement by ablating each separately, as shown in \cref{ablation_2}. Both modules contributed significantly to the final performance. In particular, the performance boost from OGM confirms that modeling future interactions enhances prediction quality, underscoring the importance of intention reasoning for improving trajectory predictions.

\noindent \textbf{Effects of Components in Mamba-Based Decoder.} We conducted an ablation analysis of various decoder components to examine the benefits of the Mamba-like structure over conventional approaches. This analysis allowed us to determine whether the design offers meaningful feature extraction enhancements or constitutes over-engineering for trajectory decoding. Results in \cref{ablation_3} highlight the advantages of this design. Compared to using MLPs as regression and classification heads, both the Bi-Mamba architecture and the self-attention mechanism between different modes significantly improved prediction performance and confidence. Additionally, we investigated the effect of the proposed dual mode tokens for classification, comparing it to a unidirectional Mamba model with a single mode token for aggregating trajectory query features. As shown in \cref{ablation_4}, the Bi-Mamba model demonstrated better performance, benefiting from its forward-backward scanning mechanism, which effectively fuses trajectory features into the two \( \text{CLS} \) tokens and validates the benefits of this design. We also examined the effects of different Mamba layer depths, as shown in \cref{ablation_5}. 
The results indicate that deeper layers can introduce unnecessary computational overhead and may also degrade performance due to overfitting, highlighting the importance of an optimal layer configuration for achieving strong performance.

\subsection{Qualitative Results}
We present visualizations of our proposed approach across a variety of traffic scenarios from the Argoverse validation set, as shown in \cref{Fig4}. These qualitative results emphasize the strong capability of our model to generate accurate, feasible, and multimodal future trajectories that remain well-aligned with the scene layout under diverse conditions, including complex intersections and long-range forecasting scenarios.  More qualitative results are provided in Appx. 7.

%% file: sec/5_conclusion.tex
\section{Conclusion}
\label{sec:5_conclusion}
In this work, we reconceptualize trajectory prediction tasks from a planning perspective and advocate a ``First Reasoning, Then Forecasting” strategy. We propose a novel and interpretable reward-driven intention reasoner, designed within a QIRL framework that combines the MaxEnt IRL paradigm with vectorized context representations through a query-centric pipeline, effectively providing informative intention priors for subsequent trajectory generation. In addition, we introduce a hierarchical DETR-like trajectory decoder integrated with a Bi-Mamba structure, which captures sequential trajectory dependencies and significantly enhances prediction accuracy and confidence. Experimental results demonstrate that our reasoning-enhanced predictor exhibits a strong capability for generating confident and reliable future trajectories that align well with the scene layout while achieving highly competitive performance against existing state-of-the-art models. Furthermore, our work underscores the critical role of intention reasoning in motion forecasting, substantiates the feasibility of RL paradigms in modeling driving behaviors, and establishes a promising baseline model for future research in trajectory prediction.

\section*{Acknowledgment}
This work was supported in part by the Hong Kong Ph.D. Fellowship Scheme, and in part by the HKUST-DJI Joint Innovation Laboratory.

%% file: main.bbl
\begin{thebibliography}{48}
\providecommand{\natexlab}[1]{#1}
\providecommand{\url}[1]{\texttt{#1}}
\expandafter\ifx\csname urlstyle\endcsname\relax
  \providecommand{\doi}[1]{doi: #1}\else
  \providecommand{\doi}{doi: \begingroup \urlstyle{rm}\Url}\fi

\bibitem[Caesar et~al.(2020)Caesar, Bankiti, Lang, et~al.]{caesar2020nuscenes}
Holger Caesar, Varun Bankiti, Alex~H Lang, et~al.
\newblock nuscenes: A multimodal dataset for autonomous driving.
\newblock In \emph{Proceedings of the IEEE/CVF conference on computer vision and pattern recognition}, pages 11621--11631, 2020.

\bibitem[Carion et~al.(2020)Carion, Massa, Synnaeve, Usunier, Kirillov, and Zagoruyko]{carion2020end}
Nicolas Carion, Francisco Massa, Gabriel Synnaeve, Nicolas Usunier, Alexander Kirillov, and Sergey Zagoruyko.
\newblock End-to-end object detection with transformers.
\newblock In \emph{European conference on computer vision}, pages 213--229. Springer, 2020.

\bibitem[Chai et~al.(2019)Chai, Sapp, Bansal, and Anguelov]{chai2019multipath}
Yuning Chai, Benjamin Sapp, Mayank Bansal, and Dragomir Anguelov.
\newblock Multipath: Multiple probabilistic anchor trajectory hypotheses for behavior prediction.
\newblock \emph{arXiv preprint arXiv:1910.05449}, 2019.

\bibitem[Chang et~al.(2019)Chang, Lambert, et~al.]{chang2019argoverse}
Ming-Fang Chang, John Lambert, et~al.
\newblock Argoverse: 3d tracking and forecasting with rich maps.
\newblock In \emph{Proceedings of the IEEE/CVF Conference on Computer Vision and Pattern Recognition}, pages 8748--8757, 2019.

\bibitem[Cui et~al.(2019)Cui, Radosavljevic, Chou, Lin, Nguyen, Huang, Schneider, and Djuric]{cui2019multimodal}
Henggang Cui, Vladan Radosavljevic, Fang-Chieh Chou, Tsung-Han Lin, Thi Nguyen, Tzu-Kuo Huang, Jeff Schneider, and Nemanja Djuric.
\newblock Multimodal trajectory predictions for autonomous driving using deep convolutional networks.
\newblock In \emph{2019 international conference on robotics and automation (icra)}, pages 2090--2096. IEEE, 2019.

\bibitem[Deo and Trivedi(2020)]{deo2020trajectory}
Nachiket Deo and Mohan~M Trivedi.
\newblock Trajectory forecasts in unknown environments conditioned on grid-based plans.
\newblock \emph{arXiv preprint arXiv:2001.00735}, 2020.

\bibitem[Deo et~al.(2021)Deo, Wolff, and Beijbom]{deo2022multimodal}
Nachiket Deo, Eric Wolff, and Oscar Beijbom.
\newblock Multimodal trajectory prediction conditioned on lane-graph traversals.
\newblock In \emph{Conference on Robot Learning}, pages 203--212, 2021.

\bibitem[Feng et~al.(2023)Feng, Zhou, Lin, Zhang, Xu, Zhang, Zhou, and Shen]{feng2023macformer}
Chen Feng, Hangning Zhou, Huadong Lin, Zhigang Zhang, Ziyao Xu, Chi Zhang, Boyu Zhou, and Shaojie Shen.
\newblock Macformer: Map-agent coupled transformer for real-time and robust trajectory prediction.
\newblock \emph{IEEE Robotics and Automation Letters}, 2023.

\bibitem[Feng et~al.(2024)Feng, Bahari, Amor, Zablocki, Cord, and Alahi]{feng2024unitraj}
Lan Feng, Mohammadhossein Bahari, Kaouther Messaoud~Ben Amor, {\'E}loi Zablocki, Matthieu Cord, and Alexandre Alahi.
\newblock Unitraj: A unified framework for scalable vehicle trajectory prediction.
\newblock \emph{arXiv preprint arXiv:2403.15098}, 2024.

\bibitem[Fernando et~al.(2020)Fernando, Denman, Sridharan, and Fookes]{fernando2020deep}
Tharindu Fernando, Simon Denman, Sridha Sridharan, and Clinton Fookes.
\newblock Deep inverse reinforcement learning for behavior prediction in autonomous driving: Accurate forecasts of vehicle motion.
\newblock \emph{IEEE Signal Processing Magazine}, 38\penalty0 (1):\penalty0 87--96, 2020.

\bibitem[Gao et~al.(2020)Gao, Sun, Zhao, Shen, Anguelov, Li, and Schmid]{gao2020vectornet}
Jiyang Gao, Chen Sun, Hang Zhao, Yi Shen, Dragomir Anguelov, Congcong Li, and Cordelia Schmid.
\newblock Vectornet: Encoding hd maps and agent dynamics from vectorized representation.
\newblock In \emph{Proceedings of the IEEE/CVF Conference on Computer Vision and Pattern Recognition}, pages 11525--11533, 2020.

\bibitem[Gilles et~al.(2021)Gilles, Sabatini, Tsishkou, Stanciulescu, and Moutarde]{gilles2021thomas}
Thomas Gilles, Stefano Sabatini, Dzmitry Tsishkou, Bogdan Stanciulescu, and Fabien Moutarde.
\newblock Thomas: Trajectory heatmap output with learned multi-agent sampling.
\newblock \emph{arXiv preprint arXiv:2110.06607}, 2021.

\bibitem[Gu and Dao(2023)]{gu2023mamba}
Albert Gu and Tri Dao.
\newblock Mamba: Linear-time sequence modeling with selective state spaces.
\newblock \emph{arXiv preprint arXiv:2312.00752}, 2023.

\bibitem[Gu et~al.(2021)Gu, Sun, and Zhao]{gu2021densetnt}
Junru Gu, Chen Sun, and Hang Zhao.
\newblock Densetnt: End-to-end trajectory prediction from dense goal sets.
\newblock In \emph{Proceedings of the IEEE/CVF International Conference on Computer Vision}, pages 15303--15312, 2021.

\bibitem[Guo et~al.(2025)Guo, Yang, Zhang, Song, Zhang, Xu, Zhu, Ma, Wang, Bi, et~al.]{guo2025deepseek}
Daya Guo, Dejian Yang, Haowei Zhang, Junxiao Song, Ruoyu Zhang, Runxin Xu, Qihao Zhu, Shirong Ma, Peiyi Wang, Xiao Bi, et~al.
\newblock Deepseek-r1: Incentivizing reasoning capability in llms via reinforcement learning.
\newblock \emph{arXiv preprint arXiv:2501.12948}, 2025.

\bibitem[Huang et~al.(2023)Huang, Liu, Wu, and Lv]{huang2023conditional}
Zhiyu Huang, Haochen Liu, Jingda Wu, and Chen Lv.
\newblock Conditional predictive behavior planning with inverse reinforcement learning for human-like autonomous driving.
\newblock \emph{IEEE Transactions on Intelligent Transportation Systems}, 2023.

\bibitem[Jaech et~al.(2024)Jaech, Kalai, Lerer, Richardson, El-Kishky, Low, Helyar, Madry, Beutel, Carney, et~al.]{jaech2024openai}
Aaron Jaech, Adam Kalai, Adam Lerer, Adam Richardson, Ahmed El-Kishky, Aiden Low, Alec Helyar, Aleksander Madry, Alex Beutel, Alex Carney, et~al.
\newblock Openai o1 system card.
\newblock \emph{arXiv preprint arXiv:2412.16720}, 2024.

\bibitem[Liang et~al.(2020)Liang, Yang, Hu, Chen, Liao, Feng, and Urtasun]{liang2020learning}
Ming Liang, Bin Yang, Rui Hu, Yun Chen, Renjie Liao, Song Feng, and Raquel Urtasun.
\newblock Learning lane graph representations for motion forecasting.
\newblock In \emph{European Conference on Computer Vision}, pages 541--556, 2020.

\bibitem[Lu et~al.(2022)Lu, Han, Ling, Wang, Chen, Varadarajan, and Covington]{lu2022kemp}
Qiujing Lu, Weiqiao Han, Jeffrey Ling, Minfa Wang, Haoyu Chen, Balakrishnan Varadarajan, and Paul Covington.
\newblock Kemp: Keyframe-based hierarchical end-to-end deep model for long-term trajectory prediction.
\newblock In \emph{2022 International Conference on Robotics and Automation (ICRA)}, pages 646--652. IEEE, 2022.

\bibitem[Mozaffari et~al.(2020)Mozaffari, Al-Jarrah, Dianati, Jennings, and Mouzakitis]{mozaffari2020deep}
Sajjad Mozaffari, Omar~Y Al-Jarrah, Mehrdad Dianati, Paul Jennings, and Alexandros Mouzakitis.
\newblock Deep learning-based vehicle behavior prediction for autonomous driving applications: A review.
\newblock \emph{IEEE Transactions on Intelligent Transportation Systems}, 23\penalty0 (1):\penalty0 33--47, 2020.

\bibitem[Nayakanti et~al.(2023)Nayakanti, Al-Rfou, Zhou, Goel, et~al.]{nayakanti2022wayformer}
Nigamaa Nayakanti, Rami Al-Rfou, Aurick Zhou, Kratarth Goel, et~al.
\newblock Wayformer: Motion forecasting via simple \& efficient attention networks.
\newblock In \emph{2023 International Conference on Robotics and Automation (ICRA)}, pages 2980--2987. IEEE, 2023.

\bibitem[Ng and Russell(2000)]{ng2000algorithms}
Andrew~Y Ng and Stuart Russell.
\newblock Algorithms for inverse reinforcement learning.
\newblock In \emph{International conference on machine learning}, page~2, 2000.

\bibitem[Ngiam et~al.(2021)Ngiam, Caine, Vasudevan, et~al.]{ngiam2021scene}
Jiquan Ngiam, Benjamin Caine, Vijay Vasudevan, et~al.
\newblock Scene transformer: A unified architecture for predicting multiple agent trajectories.
\newblock \emph{arXiv preprint arXiv:2106.08417}, 2021.

\bibitem[Pei et~al.(2020)Pei, Wu, and Wang]{pei2020quadruped}
Muleilan Pei, Dongping Wu, and Changhong Wang.
\newblock Quadruped robot locomotion in unknown terrain using deep reinforcement learning.
\newblock In \emph{2020 3rd International Conference on Unmanned Systems (ICUS)}, pages 517--522. IEEE, 2020.

\bibitem[Pei et~al.(2021)Pei, An, Liu, and Wang]{pei2021improved}
Muleilan Pei, Hao An, Bo Liu, and Changhong Wang.
\newblock An improved dyna-q algorithm for mobile robot path planning in unknown dynamic environment.
\newblock \emph{IEEE Transactions on Systems, Man, and Cybernetics: Systems}, 52\penalty0 (7):\penalty0 4415--4425, 2021.

\bibitem[Pei et~al.(2025{\natexlab{a}})Pei, Shan, Li, Shi, Huo, Gao, and Shen]{pei2025sept}
Muleilan Pei, Jiayao Shan, Peiliang Li, Jieqi Shi, Jing Huo, Yang Gao, and Shaojie Shen.
\newblock Sept: Standard-definition map enhanced scene perception and topology reasoning for autonomous driving.
\newblock \emph{IEEE Robotics and Automation Letters}, 10\penalty0 (7):\penalty0 7126--7133, 2025{\natexlab{a}}.

\bibitem[Pei et~al.(2025{\natexlab{b}})Pei, Shi, Zhang, Li, and Shen]{pei2025goirl}
Muleilan Pei, Shaoshuai Shi, Lu Zhang, Peiliang Li, and Shaojie Shen.
\newblock Goirl: Graph-oriented inverse reinforcement learning for multimodal trajectory prediction.
\newblock \emph{International Conference on Machine Learning (ICML)}, 2025{\natexlab{b}}.

\bibitem[Qi et~al.(2017)Qi, Su, Mo, and Guibas]{qi2017pointnet}
Charles~R Qi, Hao Su, Kaichun Mo, and Leonidas~J Guibas.
\newblock Pointnet: Deep learning on point sets for 3d classification and segmentation.
\newblock In \emph{Proceedings of the IEEE conference on computer vision and pattern recognition}, pages 652--660, 2017.

\bibitem[Ridel et~al.(2020)Ridel, Deo, Wolf, and Trivedi]{ridel2020scene}
Daniela Ridel, Nachiket Deo, Denis Wolf, and Mohan Trivedi.
\newblock Scene compliant trajectory forecast with agent-centric spatio-temporal grids.
\newblock \emph{IEEE Robotics and Automation Letters}, 5\penalty0 (2):\penalty0 2816--2823, 2020.

\bibitem[Ronneberger et~al.(2015)Ronneberger, Fischer, and Brox]{ronneberger2015u}
Olaf Ronneberger, Philipp Fischer, and Thomas Brox.
\newblock U-net: Convolutional networks for biomedical image segmentation.
\newblock In \emph{International Conference on Medical image computing and computer-assisted intervention}, pages 234--241. Springer, 2015.

\bibitem[Russell(1998)]{russell1998learning}
Stuart Russell.
\newblock Learning agents for uncertain environments.
\newblock In \emph{Proceedings of the eleventh annual conference on Computational learning theory}, pages 101--103, 1998.

\bibitem[Shi et~al.(2022)Shi, Jiang, Dai, and Schiele]{shi2022motion}
Shaoshuai Shi, Li Jiang, Dengxin Dai, and Bernt Schiele.
\newblock Motion transformer with global intention localization and local movement refinement.
\newblock \emph{Advances in Neural Information Processing Systems}, 35:\penalty0 6531--6543, 2022.

\bibitem[Song et~al.(2020)Song, Ding, Chen, Shen, Wang, and Chen]{song2020pip}
Haoran Song, Wenchao Ding, Yuxuan Chen, Shaojie Shen, Michael~Yu Wang, and Qifeng Chen.
\newblock Pip: Planning-informed trajectory prediction for autonomous driving.
\newblock In \emph{Computer Vision--ECCV 2020: 16th European Conference, Glasgow, UK, August 23--28, 2020, Proceedings, Part XXI 16}, pages 598--614. Springer, 2020.

\bibitem[Song et~al.(2021)Song, Luan, Ding, Wang, and Chen]{song2021learning}
Haoran Song, Di Luan, Wenchao Ding, Michael~Y Wang, and Qifeng Chen.
\newblock Learning to predict vehicle trajectories with model-based planning.
\newblock In \emph{Conference on Robot Learning}, pages 1035--1045, 2021.

\bibitem[Sutton et~al.(1998)Sutton, Barto, et~al.]{sutton1998reinforcement}
Richard~S Sutton, Andrew~G Barto, et~al.
\newblock \emph{Reinforcement learning: An introduction}.
\newblock MIT press Cambridge, 1998.

\bibitem[Tang et~al.(2024)Tang, Kan, Shan, Ji, Bai, and Chen]{tang2024hpnet}
Xiaolong Tang, Meina Kan, Shiguang Shan, Zhilong Ji, Jinfeng Bai, and Xilin Chen.
\newblock Hpnet: Dynamic trajectory forecasting with historical prediction attention.
\newblock In \emph{Proceedings of the IEEE/CVF Conference on Computer Vision and Pattern Recognition}, pages 15261--15270, 2024.

\bibitem[Varadarajan et~al.(2022)Varadarajan, Hefny, Srivastava, Refaat, Nayakanti, et~al.]{varadarajan2022multipath++}
Balakrishnan Varadarajan, Ahmed Hefny, Avikalp Srivastava, Khaled~S Refaat, Nigamaa Nayakanti, et~al.
\newblock Multipath++: Efficient information fusion and trajectory aggregation for behavior prediction.
\newblock In \emph{2022 International Conference on Robotics and Automation (ICRA)}, pages 7814--7821. IEEE, 2022.

\bibitem[Wilson et~al.(2023)Wilson, Qi, Agarwal, Lambert, Singh, Khandelwal, Pan, Kumar, Hartnett, Pontes, et~al.]{wilson2023argoverse}
Benjamin Wilson, William Qi, Tanmay Agarwal, John Lambert, Jagjeet Singh, Siddhesh Khandelwal, Bowen Pan, Ratnesh Kumar, Andrew Hartnett, Jhony~Kaesemodel Pontes, et~al.
\newblock Argoverse 2: Next generation datasets for self-driving perception and forecasting.
\newblock \emph{arXiv preprint arXiv:2301.00493}, 2023.

\bibitem[Wu et~al.(2020)Wu, Sun, Zhan, Yang, and Tomizuka]{wu2020efficient}
Zheng Wu, Liting Sun, Wei Zhan, Chenyu Yang, and Masayoshi Tomizuka.
\newblock Efficient sampling-based maximum entropy inverse reinforcement learning with application to autonomous driving.
\newblock \emph{IEEE Robotics and Automation Letters}, 5\penalty0 (4):\penalty0 5355--5362, 2020.

\bibitem[Wulfmeier et~al.(2017)Wulfmeier, Rao, Wang, Ondruska, and Posner]{wulfmeier2017large}
Markus Wulfmeier, Dushyant Rao, Dominic~Zeng Wang, Peter Ondruska, and Ingmar Posner.
\newblock Large-scale cost function learning for path planning using deep inverse reinforcement learning.
\newblock \emph{The International Journal of Robotics Research}, 36\penalty0 (10):\penalty0 1073--1087, 2017.

\bibitem[Yao et~al.(2023)Yao, Li, Lang, and Chuah]{yao2023goal}
Zhen Yao, Xin Li, Bo Lang, and Mooi~Choo Chuah.
\newblock Goal-lbp: Goal-based local behavior guided trajectory prediction for autonomous driving.
\newblock \emph{IEEE Transactions on Intelligent Transportation Systems}, 2023.

\bibitem[Zhang et~al.(2024{\natexlab{a}})Zhang, Song, and Zhang]{zhang2024decoupling}
Bozhou Zhang, Nan Song, and Li Zhang.
\newblock Decoupling motion forecasting into directional intentions and dynamic states.
\newblock \emph{arXiv preprint arXiv:2410.05982}, 2024{\natexlab{a}}.

\bibitem[Zhang et~al.(2022)Zhang, Li, Chen, and Shen]{zhang2022trajectory}
Lu Zhang, Peiliang Li, Jing Chen, and Shaojie Shen.
\newblock Trajectory prediction with graph-based dual-scale context fusion.
\newblock In \emph{2022 IEEE/RSJ International Conference on Intelligent Robots and Systems (IROS)}, pages 11374--11381. IEEE, 2022.

\bibitem[Zhang et~al.(2024{\natexlab{b}})Zhang, Li, Liu, and Shen]{zhang2024simpl}
Lu Zhang, Peiliang Li, Sikang Liu, and Shaojie Shen.
\newblock Simpl: A simple and efficient multi-agent motion prediction baseline for autonomous driving.
\newblock \emph{IEEE Robotics and Automation Letters}, 2024{\natexlab{b}}.

\bibitem[Zhou et~al.(2022)Zhou, Ye, Wang, Wu, and Lu]{zhou2022hivt}
Zikang Zhou, Luyao Ye, Jianping Wang, Kui Wu, and Kejie Lu.
\newblock Hivt: Hierarchical vector transformer for multi-agent motion prediction.
\newblock In \emph{Proceedings of the IEEE/CVF Conference on Computer Vision and Pattern Recognition}, pages 8823--8833, 2022.

\bibitem[Zhou et~al.(2023)Zhou, Wang, Li, and Huang]{zhou2023query}
Zikang Zhou, Jianping Wang, Yung-Hui Li, and Yu-Kai Huang.
\newblock Query-centric trajectory prediction.
\newblock In \emph{Proceedings of the IEEE/CVF Conference on Computer Vision and Pattern Recognition}, pages 17863--17873, 2023.

\bibitem[Zhu et~al.(2024)Zhu, Liao, Zhang, Wang, Liu, and Wang]{zhu2024vision}
Lianghui Zhu, Bencheng Liao, Qian Zhang, Xinlong Wang, Wenyu Liu, and Xinggang Wang.
\newblock Vision mamba: Efficient visual representation learning with bidirectional state space model.
\newblock \emph{arXiv preprint arXiv:2401.09417}, 2024.

\bibitem[Ziebart et~al.(2008)Ziebart, Maas, Bagnell, Dey, et~al.]{ziebart2008maximum}
Brian~D Ziebart, Andrew~L Maas, J~Andrew Bagnell, Anind~K Dey, et~al.
\newblock Maximum entropy inverse reinforcement learning.
\newblock In \emph{AAAI}, pages 1433--1438. Chicago, IL, USA, 2008.

\end{thebibliography}
